# ADVANCED TRANSIENT DIAGNOSTIC WITH ENSEMBLE DIGITAL TWIN MODELING

**Edward Chen[1], Linyu Lin[2], and Nam T. Dinh[3]**

Department of Nuclear Engineering, North Carolina State University Raleigh, NC, 27695, USA

[1]echen2@ncsu.edu, [2]llin7@ncsu.edu, [3]ntdinh@ncsu.edu

## Abstract

The use of machine learning (ML) model as digital-twins for reduced-order-modeling (ROM) in lieu of system codes has grown traction over the past few years. However, due to the complex and non-linear nature of nuclear reactor transients as well as the large range of tasks required, it is infeasible for a single ML model to generalize across all tasks. In this paper, we incorporate issue specific digital-twin ML models with ensembles to enhance the prediction outcome. The ensemble also utilizes an indirect probabilistic tracking method of surrogate state variables to produce accurate predictions of unobservable safety goals. The unique method named Ensemble Diagnostic Digital-twin Modeling (EDDM) can select not only the most appropriate predictions from the incorporated diagnostic digital-twin models but can also reduce generalization error associated with training as opposed to single models.

## Introduction

In recent years, autonomous control systems have been encouraged in advanced reactors to help simplify operations and maintenance, and enable remote-site power generation. The nearly autonomous management and control (NAMAC) system is designed to provide effective action recommendations to the operator for maintaining the safety and performance under specified goals. Recommendations are made based on NAMAC's knowledge of the current plant state, prediction of future state transients, and reflection on known uncertainties that complicate the determination of mitigating strategies. Such knowledge is extracted by machine-learning (ML) algorithms and stored as digital-twins (DTs) of various sub-functions. For example, the diagnosis DT (D-DT) portion monitors and reports safety-significant factors based on incoming real-time sensory data. This information is then relayed to prognostic and strategy management DTs which, based on consequences and user-defined goals, generate appropriate course-of-action recommendations [1].

For diagnostic purposes, the aim of the D-DT is to assimilate data from the plant and then evaluate the state of the physical system, including unobservable variables, operating, and fault conditions. D-DT are a type of fault detection and diagnosis (FDD) models. Here, model-free FDDs (especially data-driven methods) have an advantage over traditional model-based FDDs as they are not explicitly reliant on mathematical models. By constructing D-DT with ML approaches, for instance artificial neural networks (ANNs), any non-linear relationship between correlated system states can be achieved with any desired degree of accuracy. Significant progress has already been made with data-driven model-free FDDs.

In this study, D-DTs are used to determine the unobservable safety-significant state variables, defined as Safety-Significant Factors (SSFs) and can include peak fuel centerline and cladding temperatures. Meanwhile, NAMAC will make recommendations based on these parameters, which are related to reactor performance and safety. As NAMAC is expected to make reliable decisions for various scenarios, the DDT is required to have high accuracy for all designed issue spaces. However, the development of all-encompassing DTs typically requires very deep neural networks and long training times [2]. In addition, singular supervised models can be constrained due to bias-variance trade-off associated with over-generalization. With that respect, a framework is needed to extract and organize the copious amounts of data into accurate and meaningful metrics (i.e., SSFs) of the unfolding transient while also maintaining low DT generalization error.

A well-documented technique that offers both versatility and accuracy are ensembles; simply a collection of models where each output is aggregated together based on the relative 'correctness' of each prediction (i.e., weighted average, error-based discrepancy, etc.). By combing the predictions of multiple neural networks, the resultant output is less sensitive to the specifics associated with training (i.e., choice of training scheme, stochastically asymptotic single runs, etc.) ergo reducing over-fitting related errors [3]. In addition, for a given problem, there may be more than one learner that performs equally well, thus through combination, the overall prediction is more informed than any single learner [4]. Effective adaptive ensembles can also be created from libraries of pre-trained models, providing a non-rigid prediction framework [5]. In this study, ensembles of data-driven DTs will be used to predict SSFs in a simulated transient scenario.

## Objective

To demonstrate the capability of Ensemble Diagnostic Digital-twin Modeling (EDDM), partial loss-of-flow accident scenarios are simulated using a GOTHIC model of the Experimental Breeder Reactor-II (EBR-II) [6]. In EBR-II, two separate primary sodium pumps (designated PSP#1 and #2) provide coolant flow through the core [7]. In the postulated scenario, PSP#1 partially losses pump rotational speed, decreasing the overall coolant flow through the core block. The scenario is detected and monitored by the DDT and based on NAMAC's recommendation, the rotational speed of PSP#2 increases to compensate. The objective is to preserve reactor safety by keeping the SSFs below a prescribed limit. Without a recommendation by NAMAC or if the scenario is undetected, the decrease in overall core flow rate will result in the fuel centerline temperature ($T_{FCL}$) exceeding safe operational values and core damage. As such, $T_{FCL}$ is designated as an SSF and will be predicted by the EDDM. However, as $T_{FCL}$ is generally an unobservable variable due to the sensor proximity to obstructions within the fuel bundle, the error associated with individual DT $T_{FCL}$ predictions cannot be calculated and thus cannot be used to aggregate the predictions together. Instead, an indirect approach must be adopted using surrogate tracking parameters that are correlated to and indicative of $T_{FCL}$. For this transient, the upper plenum temperature and total core flow rate are the surrogate tracking parameters and are chosen based on the strong correlation to $T_{FCL}$ calculated via the Pearson correlation coefficient.

The scope of the numerical demonstration can be fully represented by the time-dependent curve of the rotational speed, $w_1(t)$, of PSP1 defined by Eqn. (1.

$$w_1(t) = w_0 \left(1 - \frac{1 - (w_1)_{end}}{T_1} t \right), \qquad t_{acc} \leq t \leq t_{acc} + T_1 \tag{1}$$

Here, $w_0$ is the nominal pump speed, $T_1$ is the ramp-down duration, $(w_1)_{end}$ is the normalized PSP#1 end speed, and $t_{acc}$ marks the time when the transient begins. By varying the pump end speed and ramp down duration, different ramp down profiles can be achieved. In total 15 different measurements are sampled for 2000 time-steps per episode, notably the core flow rate, plenum temperatures, and the unobservable $T_{FCL}$. The objective of the ensemble is as follows: given a set of sensor measurements from the simulation, choose indirectly via surrogate predicted tracking parameters and aggregate the best $T_{FCL}$ predictions.

## Methodology

There are three primary components to an ensemble typically used in literature: feature variation, dataset selection, and voting aggregation. Feature variation tackles the problem where different models have different predicative capabilities based on which inputs are provided during training. This is due to the varying correlation strengths between inputs and outputs and significantly affects model fidelity. In a dataset with numerous input features, the task of determining the appropriate features can be intractable. Typical ensembles deal with this issue by generating large number of models with random feature selection ensuring at least some will converge on the optimal configuration. However, the associated number of models required to be effective scale exponentially with the number of inputs making this method highly inefficient. Instead, more conventional methods are used for this paper, namely the Pearson correlation coefficient between the simulated sensor values ($X$) and the fuel centerline temperature ($Y$) (Eqn. (2). Additional inputs are also selected based on expert experience about the scenario.

$$\rho_{X,Y} = \frac{cov(X,Y)}{\sigma_X \sigma_Y} \tag{2}$$

In **Figure 1**, the layout of the EDDM scope is presented within the larger NAMAC operational framework. The EDDM is composed of N parallel D-DT, all predicting the same SSF using sensory data received from the plant simulator. The senor data is also sent to the Probabilistic Voting Aggregation (PVA) block that decides how to aggregate the predictions together based on relevant contextual information. Specifically, it calculates the conditional probabilities of each DT based on the associated tracking error. The output of the EDDM is the aggregated SSF prediction which is passed forward to prognosis and strategy management for additional analysis.

The D-DT models are 3-layer feed forward neural networks (FNN) with ReLu activation functions [8]. Training of the separate DTs are based on the severity of pump degradation calculated from Eqn. (1. In this case, three distinct datasets are formed labelled Train, Intp., and Extp. for training, interpolation and extrapolation tasks. The specifications can be found in Table 1. It is assumed during training that the SSF and the surrogate tracking parameters are known and can be minimized using back propagation. Both k-fold cross validation [9] and L2 regularization [10] are used to manage overfitting of the FNN networks.

The trained FNN models are then stored as DTs for that transient scenario within a library and can be later retrieved by the EDDM.

After each of the base models are trained, the outputs are aggregated together via the PVA block. Conventional methods include weighted averaging [5], where the weights reflect the relative performance of each individual model. However, determining the weights is a non-trivial task and can involve grid searching on an additional validation dataset. Furthermore, depending on the data coverage of the validation set, the weights can be biased towards a specific scenario and underrepresent the true test space. To overcome this issue, the output of the EDDM borrows an idea from control theory to establish accurate indirect tracking of the $T_{FCL}$ via surrogate parameters. A PID controller (proportional, integral, derivative) is implemented to calculate the total error between the tracking parameters predicted by each individual DT and the measurable true tracking parameters from the simulator (Eqn. (3)). The models can then be selected based on the relative error of the surrogate parameters.

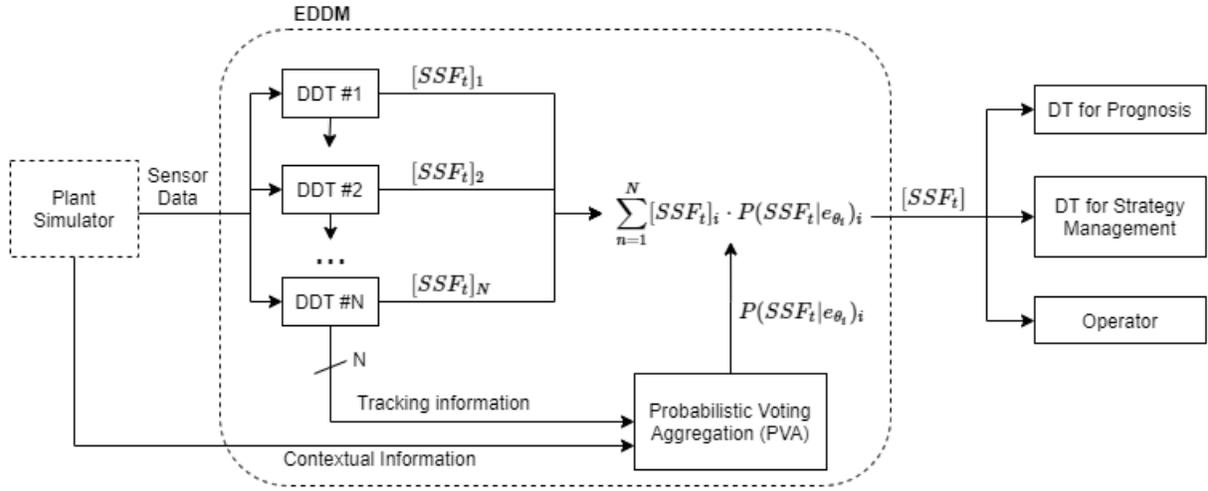

Figure 1. Schematic of EDDM framework within larger NAMAC scope

| Sample Size | $T_1$ | $(w_1)_{end}$ (% of $w_0$) | Identification |
|---|---|---|---|
| 1024 | 467.81 | 51.6 – 100 | Train |
| 250 | 467.81 | 51.6 – 100 | Intp. |
| 250 | 467.81 | 0 – 38.7 | Extp. |

Table 1. Training and Testing Datasets for models within EDDM

$$e_k(\vec{\theta}_{k,t_i}, t_i) = a \cdot e_P(\vec{\theta}_{k,t_i}, t_i) + b \cdot e_I(\vec{\theta}_{k,t_i}, t_i) + c \cdot e_D(\vec{\theta}_{k,t_i}, t_i) \qquad (3)$$

$$e_P(\vec{\theta}_{k,t_i}, t_i) = \omega^T[\vec{\theta}_{t_i} - \vec{\theta}_{k,t_i}] \qquad (4)$$

$$e_I(\vec{\theta}_{k,t_i}, t_i) = \omega^T\left[\int_{t=0}^{t_i}(\vec{\theta}_{t_i} - \vec{\theta}_{k,t_i})dt\right] \qquad (5)$$

$$e_D(\vec{\theta}_{k,t_i}, t_i) = \omega^T\left[\frac{d\vec{\theta}_{t_i}}{dt} - \frac{d\vec{\theta}_{k,t_i}}{dt}\right] \qquad (6)$$

Here, $\vec{\theta}_{k,t_i}$ is the predicted tracking parameters at the $t_i$ time step by the $k^{th}$ model, $\vec{\theta}_{t_i}$ is the measured tracking parameter, and $e_k$ is the scalar total error associated with the $k^{th}$ model. The coefficients $a$, $b$, and $c$ denote the relative importance of offset, cumulative, and trend error through the transient. The weight matrix $\omega^T$ on the other hand, denotes the importance of each surrogate tracking parameter. If only one of the models' predictions in the ensemble are used, this would be highly computationally inefficient. Instead, all relevant overlapping DT predictions can be utilized by calculating the conditional prediction likelihood based on the tracking parameter error via the negative log likelihood of the normalized error [11] (Eqn. (7). As the tracking parameters are highly correlated to and indicative of the SSF, a lower tracking error is suggestive of more accurate SSF predictions. This assumption is only true when the included DDTs are approximately representative of the transient [12]. The EDDM output after voting aggregation is thus the sum product of the prediction probabilities with the predicted SSF of each model (Eqn. (3).

$$P\left(\hat{y}_{k,t_i} \middle| e_k(\vec{\theta}_{k,t_i}, t_i)\right) = -\log[|e_k(\vec{\theta}_{k,t_i}, t_i)|] \tag{7}$$

$$\hat{y}_{t_i} = \sum_{\{k=1\}}^{K} P\left(\hat{y}_{k,t_i} \middle| e_k(\vec{\theta}_{k,t_i}, t_i)\right) \cdot \hat{y}_{k,t_i} \tag{8}$$

## Results

In this section, the verification and validation of the EDDM is completed. The test issue space is described in Eqn. (1. For testing, 18 separate D-DT FNNs are developed to act as the base models for the EDDM. Testing was conducted on the Intp. and Extp. dataset and include transients with pump degradation similar to and more severe as those seen during the training phase. The tracked surrogate parameters include the upper plenum temperature and the total core flow rate with default PID coefficients and weight matrix values of [10, 0.5, 0.8] and [0.5, 0.8]. To gauge the general performance of the models, the mean square error between the EDDM prediction and the true SSF value is calculated. As a visual example of EDDM capability in comparison to a single FNN model, a randomly selected LOFA episode is predicted (**Figure 2**). The mean square error is calculated for the entire transient.

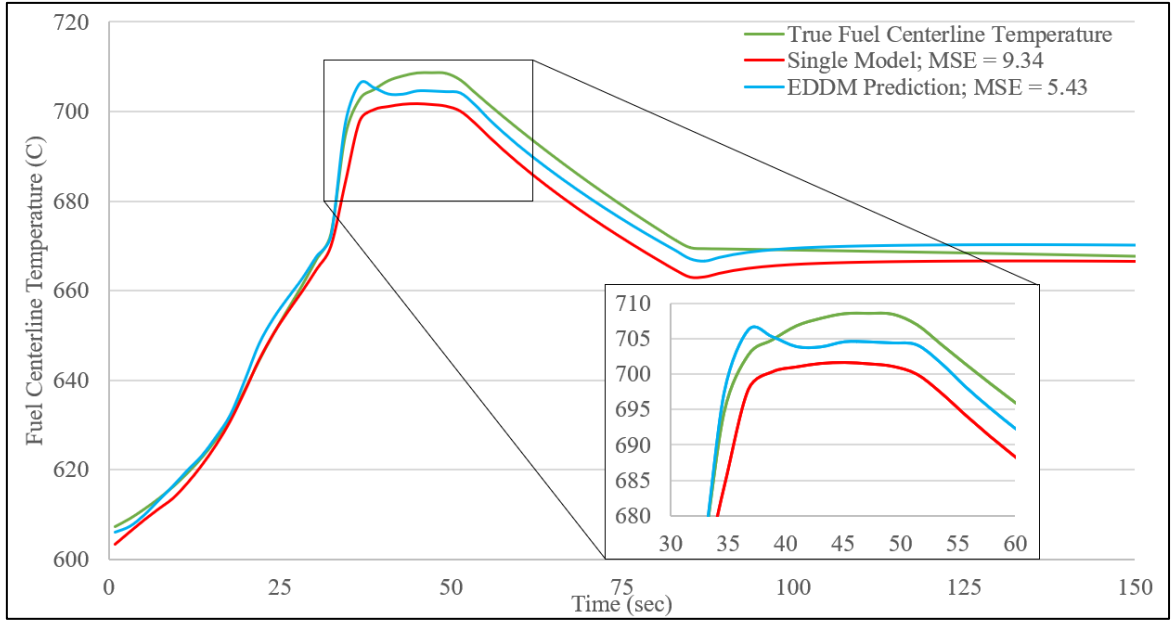

**Figure 2. Example of transient prediction made by EDDM and single generalized model for randomly selected LOFA scenario**

For comprehensive testing of the EDDM, testing and training conditions are varied and compared to single models with varying data coverage (**Table 2**). The testing conditions and database details can be found in Table 1. In the left column, the different testing methods are presented, where the percent represents the number of the training samples in relation to the training database(s). In the single model cases, error bars are generated as the final training MSE varied about the target MSE of 2.5. Therefore, 25 single models are trained and used to predict on the same dataset with the mean and 95% capture boundaries of the outcome included into the table. Coincidentally, this further illustrates the necessity for ensembles to compensate for training related bias and variance uncertainties. The results for the Intp. and Extp. dataset can be seen in the last two columns. Different data coverage conditions are simulated to examine the over generalization problem for single models.

|  | Case # | Intp. MSE ($^oC^2$) | Extp. ($^oC^2$) |
|---|---|---|---|
| EDDM V2 (Train 5%) | 1 | 2.5 | 6.2 |
| EDDM V3 (Train & Extp. 5%) | 2 | 2.6 | 4.9* |
| Single (Train 5%) | 3 | 4.0 ± 0.23 | 8.6 ± 0.70 |
| Single (Train & Extp. 5%) | 4 | 3.9 ± 0.13 | 7.9* ± 0.13 |
| Single (Train & Extp. 10%) | 5 | 4.0 ± 0.10 | 8.0* ± 0.10 |
| Single (Train & Extp. 40%) | 6 | 3.2 ± 0.05 | 6.4* ± 0.05 |

*Not true extrapolation, tasks are treated as interpolation even with partial coverage.

**Table 2. Case study of EDDM performance comparison with single models under different data coverage conditions**

Comparing case 1 and case 3, it is immediate evident that the EDDM outperforms the overgeneralized single FNN model in both the interpolation and extrapolation scenarios. Furthermore, only with significant data coverage of the single model can it perform at a level equal to the EDDM (case 3 - 6). If base models with some coverage of the extrapolated scenarios are included into the EDDM, the performance improves without significantly degrading interpolation capabilities (case 2). The key benefit from this is the ability to expand the DT-D library with new models as more relevant data is available without compromising existing models already in the repository.

Examining the individual base models' MSE (**Figure 3**) within the EDDM for a randomly selected episode, it is evident not every DT performs well. Instead, only models with training sets comparable to the test transient perform well. Models that have disjoint training sets relative to the test will exhibit overall higher mean square error for the entire transient. Through use of the PVA, the EDDM output has an MSE approximately equivalent to the best base models for the entire transient and suggests successful indirect PID tracking of the SSFs.

In **Table 3**, the EDDM is tested again on interpolation and extrapolation tasks. However, in addition to mean square error, the largest and smallest one-time temperature excursion is also determined. This value is important in dictating the control action sent to the operator and indicates time instantaneous false positive or negative scenarios. For example, should the EDDM under predict the fuel centerline temperature by more than $10^oC$, the true damage to the fuel could be moderate to severe. The prediction results demonstrate capability in avoiding large instantaneous temperature excursions. This can be further deduced from **Figure 3**, where while any individual model within the EDDM can have large prediction errors, it does not necessarily manifest in the final EDDM output. Lastly, the excursions are within a $\pm 10^oC$ limit and is a positive indication that the ensemble can reduce the likelihood for unnecessary SCRAM actions.

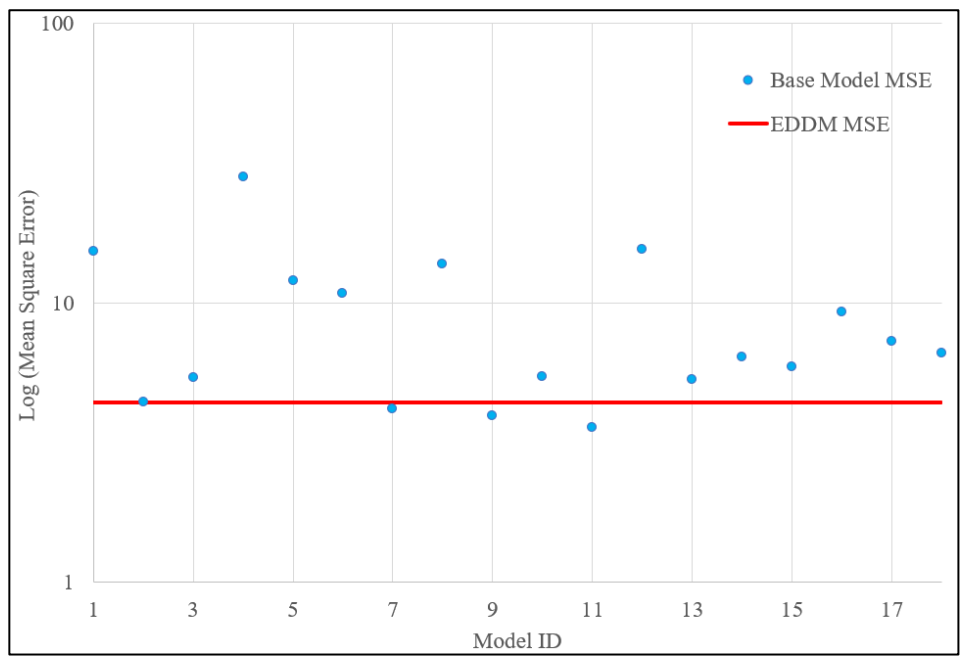

Figure 3. Example of base model MSE compared to EDDM total MSE for a randomly selected LOFA scenario.

| Dataset | Sample Count | EDDM Prediction MSE | Largest Positive Temp. Excursion | Largest Negative Temp. Excursion |
|---|---|---|---|---|
| Intp. | 250 | 2.71 | 10.29 | -7.92 |
| Extp. | 250 | 6.21 | 8.15 | -7.03 |

Table 3. Interpolation and extrapolation capability of EDDM with largest and smallest temperature excursions

From a global perspective, the MSE is an acceptable metric when comparing between complete transients, however, does not inform on local phenomenon that can occur during a transient. Such local phenomenon can include additional control actions or crude break off that nearly instantaneously change the transient behavior. These scenarios are referred as context switches and can result in large bias errors. In **Figure 4**, to test the impact of switches on EDDM, a second control action is injected at $T = 120$ during an ongoing transient. While the EDDM is still capable of tracking the SSFs within the acceptable margin, there is no notification of a context switch or that the transient has changed from one regime to another. Here, individual models that perform well initially will have growing errors after the switch. However, this detection of local phenomenon is obscured due to the aggregation of base model results.

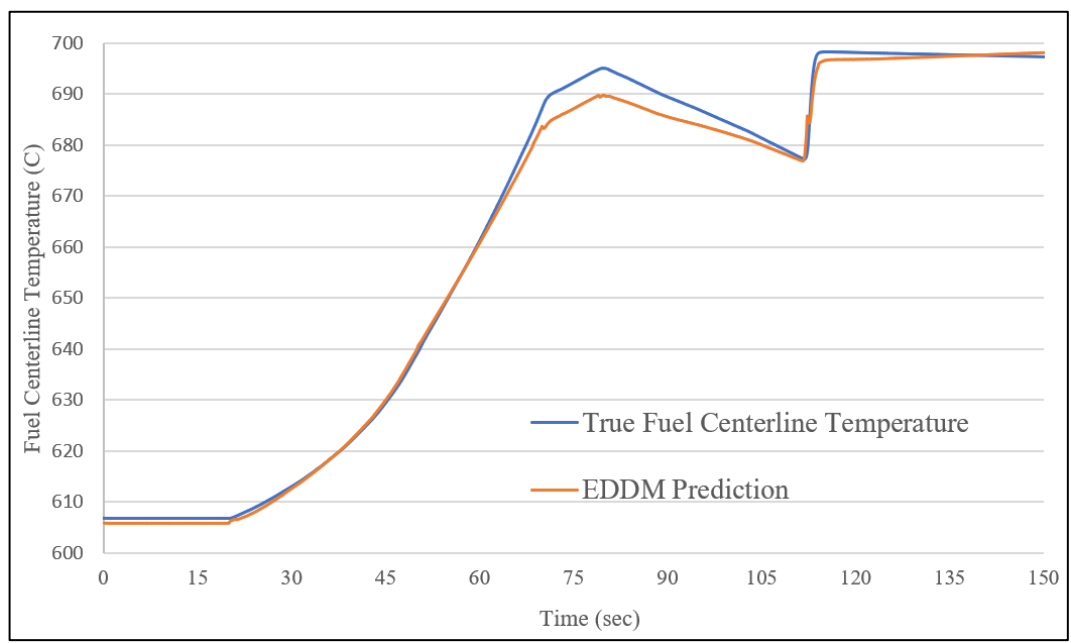

Figure 4. Example of multiple control actions during an ongoing transient with EDDM response

# Conclusion

In this paper, a novel method for indirect transient diagnostics is presented using ML-based DTs within an ensemble framework. The inclusion of the ensemble methods can meaningfully reduce generalization error in comparison to single models. In addition, the use of PVA and PID tracking of surrogate parameters allows accurate predictions of unobservable SSFs and represents a new approach to real-time transient diagnostics. Future work is dedicated to detecting local phenomenon and contextual switches in machine learning diagnostics ensembles. In addition, ongoing work is being conducted to expand the D-DT library to include transients of different accident scenarios.

## Acknowledgements

This work is performed with the support of the ARPA-E MEITNER program under the project titled: 'Development of a Nearly Autonomous Management and Control System for Advanced Reactors'. The authors would like to thank Dr. Pascal Rouxelin of North Carolina State University for data generation and management.

## References

[1] L. Lin, P. Athe, P. Rouxelin, R. Youngblood, A. Gupta, J. Lane, M. Avramova and N. Dinh, "Development and Assessment of a Nearly Autonomous Management and Control System for Advanced Reactors," 29 August 2020. [Online]. Available: https://arxiv.org/abs/2008.12888. [Accessed 7 Oct 2021].

[2] H. W. Lin, M. Tegmark and D. Rolnick, "Why does deep and cheap learning work so well?," *Journal of Statistical Physics,* vol. 168, no. 6, pp. 1223-1247, 2017.

[3] T. G. Dietterich, "Machine-learning research," *AI magazine,* vol. 18, no. 4, p. 97, 1997.

[4] A. Lacoste, M. Marchand, F. Laviolette and H. Larochelle, "Agnostic Bayesian learning of ensembles," *Proceedings of the International Conference on Machine Learning,* pp. 611-619, 2014.

[5] R. Caruana, A. Niculescu-Mizil, G. Crew and A. Ksikes, "Ensemble selection from libraries of models," *Proceedings of the twenty-first international conference on Machine learning,* p. 18, 2004.

[6] J. Lane, J. M. Link, J. M. King, T. L. George and S. W. Claybrook, "Benchmark of GOTHIC to EBR-II SHRT-17 and SHRT-45R Tests," *Nuclear Technology,* vol. 206, no. 7, pp. 1019-1035, 2020.

[7] S. A. Kamal and D. Hill, "Fault Tree Analysis of the EBR-II Reactor Shutdown System," *Probabilistic safety assessment international topical meeting,* vol. 24, no. 1, 1993.

[8] T. L. Fine, Feedforward neural network methodology, Berlin: Springer Science & Business Media, 2006.

[9] P. Refaeilzadeh, L. Tang and H. Liu, "Cross-Validation," in *LIU L., OZSU M.T. (eds) Encyclopedia of Database Systems*, Berlin, Springer, 2009.

[10] A. Y. Ng, "Feature selection, L1 vs. L2 regularization, and rotational invariance," in *Proceedings of the twenty-first international conference on Machine learning*, 2004.

[11] P. A. Bosman and D. Thierens, "Negative log-likelihood and statistical hypothesis testing as the basis of model selection in IDEAs," 2000.

[12] R. Caruana, A. Munson and A. Niculescu-Mizil, "Getting the most out of ensemble selection," in *Sixth International Conference on Data Mining (ICDM'06)*, 2006.

[13] B. Hanin, "Universal Function Approximation by Deep Neural Nets with Bounded Width and ReLU Activations," *Mathematics,* vol. 992, no. 7, 2019.